**Comprehensive Benchmarking of Machine Learning Methods for Risk Prediction Modelling from Large-Scale Survival Data: A UK Biobank Study**

[1]Oexner, RR*; [1]Schmitt, R*; [1]Ahn, H*; [2]Shah, RA; [1]Zoccarato, A; [1]Theofilatos, K; [1]Shah, AM

*These authors contributed equally

[1] King's College London British Heart Foundation Centre of Research Excellence, School of Cardiovascular & Metabolic Medicine and Sciences, King's College London, London, United Kingdom

[2] Institute of Cardiovascular Science, University College London, London, United Kingdom

1. Abstract

Predictive modelling is vital to guide preventive efforts. Whilst large-scale prospective cohort studies and a diverse toolkit of available machine learning (ML) algorithms have facilitated such survival task efforts, choosing the best-performing algorithm remains challenging. Benchmarking studies to date focus on relatively small-scale datasets and it is unclear how well such findings translate to large datasets that combine omics and clinical features.

We sought to benchmark eight distinct survival task implementations, ranging from linear to deep learning (DL) models, within the large-scale prospective cohort study UK Biobank (UKB). We compared discrimination and computational requirements across heterogenous predictor matrices and endpoints. Finally, we assessed how well different architectures scale with sample sizes ranging from $n$ = 5,000 to $n$ = 250,000 individuals.

Our results show that discriminative performance across a multitude of metrics is dependent on endpoint frequency and predictor matrix properties, with very robust performance of (penalised) COX Proportional Hazards (COX-PH) models. Of note, there are certain scenarios which favour more complex frameworks, specifically if working with larger numbers of observations and relatively simple predictor matrices. The observed computational requirements were vastly different, and we provide solutions in cases where current implementations were impracticable.

In conclusion, this work delineates how optimal model choice is dependent on a variety of factors, including sample size, endpoint frequency and predictor matrix properties, thus constituting an informative resource for researchers working on similar datasets. Furthermore, we showcase how linear models still display a highly effective and scalable platform to perform risk modelling at scale and suggest that those are reported alongside non-linear ML models.

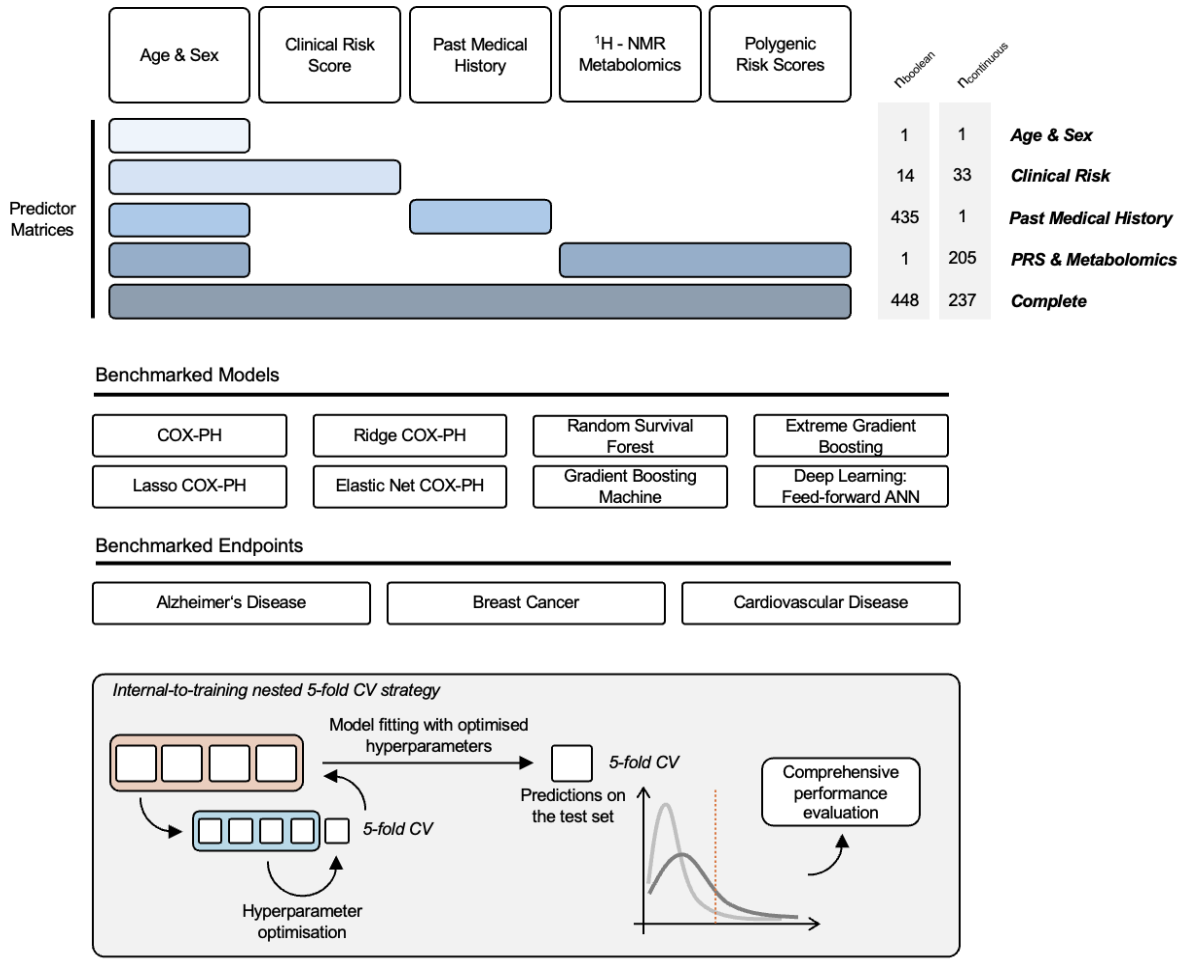

**Figure 1: Graphical Abstract.** Within this work, we test eight distinct machine learning algorithms across three disease endpoints and five heterogenous predictor matrices in a nested 5-fold cross-validation (CV) approach. We systematically assess and report discriminative performance and computational resource requirements. $^1$H-NMR: proton nuclear magnetic resonance spectroscopy, PRS: polygenic risk scores, COX-PH: COX proportional hazards, ANN: artificial neural network.

2. Introduction

Primary prevention efforts are a cost-effective solution to facilitate healthy ageing, combat multi-morbidity and free up healthcare system resources[1,2]. Within this framework, reliable and accurate risk stratification is crucial to focus attention on individuals most likely to benefit from additional monitoring or early intervention, thus improving the overall cost-effectiveness of healthcare. Consequently, the development and iterative improvement of risk scoring systems for a variety of conditions is gaining significant traction amongst the research community as well as corporate and institutional healthcare providers worldwide.

Over the past decade, large-scale biological and clinical datasets have grown in both quantity and complexity, driven by technological advances in genomics, metabolomics, proteomics, and other high throughput 'omics' fields. Large-scale prospective cohort studies (such as the UKB cohort[3,4]) offer genetic, biochemical, and lifestyle data, together with long-term longitudinal electronic health record follow-up, for hundreds of thousands of participants. Utilizing these resources, researchers now begin to dissect complex trait architectures that had previously been obscured by small sample sizes or incomplete phenotyping. Such efforts increasingly integrate a variety of ML methods[5-7], which offer flexible and increasingly accessible means of capturing complex, nonlinear relationships.

In this context, leveraging the theoretical benefits of choosing the best-performing model for a specific research question is of significant practical concern for researchers wishing to adopt state-of-the-art methods on large datasets. Much of the existing benchmarking literature[8-11] focuses on relatively small sample sizes or limited predictor matrices. These smaller-scale benchmarks have undoubtedly provided valuable insights into model performance under controlled conditions, but do not necessarily reflect the realities of big data survival tasks. Besides potentially significant differences in model performance with increasing sample size, some predictor matrices or specific hyperparameter settings might push algorithms beyond their practical limits in terms of computational requirements. Benchmarking different methods and tuning hyperparameter combinations in a per-project manner can quickly become prohibitively expensive with large datasets. Furthermore, choosing the optimal method is further complicated by the frequency of the endpoint in question. Rare endpoints can create severe class imbalance, which might complicate model training and lead to underperformance of certain methods. The outcome frequency interacts with the predictive power of the available features and can drastically shift the relative advantages of different modelling strategies. Given these interdependencies, it becomes vital to systematically compare the discriminative potential and computational burdens of a variety of analytical methods across a range of endpoint frequencies and predictor matrix properties at scale.

To address these questions, we undertook a comprehensive benchmarking study focused on large-scale survival tasks. Our aim was not only to measure and compare the discriminative power of different ML frameworks (ranging from linear models to DL architectures), but also to document their computational footprints across five highly heterogenous predictor matrix properties and three distinct endpoints. In a nested cross-validation approach, we systematically optimize each method via extensive hyperparameter tuning, rather than comparing default parameter settings that might arbitrarily favour one approach over another. We showcase that discriminative performance and compute metrics shift based on endpoint frequency and predictor matrix properties, potentially rendering some methods superior in certain scenarios and impractical in others. Moreover, we introduce and systematically benchmark a novel survival functionality for *LightGBM*[12] (as repeatedly requested by the community[13]), which offers considerably reduced computational resource requirements ($10^2$ faster than *scikit-survival*'s implementation on $n$ = 50,000 samples) over existing solutions, whilst maintaining or even improving performance.

In summary, this large computational effort - spanning multiple state-of-the-art algorithms - serves as a practical and comprehensive guide for the growing collective of researchers wishing to apply survival task ML algorithms at scale. As part of this work, we extend the toolkit by providing a robust, fast and scalable option for gradient boosting machine (GBM) – based survival analyses.

3. Methods

3.1 Study design and population

UKB is a large-scale prospective cohort study representative of the general UK population. Individuals were recruited between 2006-2010 based on (I) living in a 25-mile radius around one of 22 assessment centres scattered throughout the UK, (II) being aged 40-69 years old and (III) their capacity to consent. Approximately 5% of invited individuals ($n \approx 500{,}000$ participants) voluntarily participated in the study. Upon their initial assessment centre visit, written consent was given, all individuals filled extensive questionnaires, conducted verbal interviews, and received extensive physical measurements. Various relevant biological specimens, including blood samples were collected and stored.

UKB regularly adds and expands their resource via several routes: (I) results fed back from researchers working on the platform, (II) ongoing analysis of biological specimens (e.g. different -omics assays conducted on previously collected blood samples), (III) interval invitation of a subset of individuals, repeating aforementioned extensive phenotyping with biological sampling and (IV) ongoing follow-up via electronic health records (Hospital Episode Statistics in England, Patient Episode Database for Wales, and Scottish Morbidity Record) and death register (NHS England and NHS Central Register) linkage.[4,14]

3.2 Endpoint definition

All endpoint definitions were adapted as previously described[5,15-17]. To cover old health records, we mapped ICD-10 to ICD-9 codes in cases where possible without ambiguity (as provided by UKB, Data-Coding 1836). First occurrence of the event was defined as the earliest event record in either (I) electronic health records (which also include operation and procedure codes), (II) death records or (III) verbal interviews as part of UKB's repeat assessment visits. Detailed endpoint definitions are provided in Supplementary Table 1.

3.3 Predictor matrices

We defined five distinct predictor matrices with highly heterogenous features to assess model performance in different settings. The feature classes were restricted to either continuous or boolean (vide infra, "Preprocessing and Imputation").

(I) Age & Sex: As age and sex are freely available and highly predictive for a range of diseases, they are assumed to constitute a vital input for any risk modelling approach. We therefore assessed age and sex alone but also considered them as part of all subsequently presented predictor matrices. We included $n_{features} = 2$ ($n_{continuous} = 1$, $n_{boolean} = 1$) columns.

(II) Clinical Risk: We adapted an extensive clinical risk panel with predictive value across a multi-disease spectrum from Buergel et al.[5]. Amongst the variables is a mixture of sociodemographic and lifestyle factors, physical measurements, and a range of clinical chemistry measurements. Additional data points integrate past medical history, family history and current medication. In addition to Age & Sex, this comprised $n_{features}$ = 45 ($n_{continuous}$ = 13, $n_{boolean}$ = 32) columns.

(III) Polygenic Risk Scores (PRS) & Metabolomics: We utilized data from two high-dimensional, high-throughput -omics platforms, both of which are commonly used for risk modelling. We included proton nuclear magnetic resonance spectroscopy ($^1$H-NMR) metabolomics, conducted for UKB in cooperation with Nightingale Health Plc on non-fasting, venous ethylenediamine tetraacetic acid (EDTA) plasma samples, for a subset of n ≈ 280,000 individuals. Specifically, we leveraged all original metabolite measurements, spanning 168 data points covering heterogenous metabolic pathways with particularly good resolution for lipid-subspecies. We also utilized "standard" PRS, as returned by Thompson et al.[17], which were entirely derived from non-UKB genome-wide association study (GWAS) data and calculated/ancestry-centred for all UKB individuals. We included all available 36 PRS, including PRS for the three endpoints of interest (Alzheimer's disease, Breast Cancer, Cardiovascular Disease), irrespective of whether they were specifically constructed for or related to this studies' endpoints. In addition to Age & Sex, this comprised $n_{features}$ = 204 ($n_{continuous}$ = 204, $n_{boolean}$ = 0) columns.

(IV) Past Medical History (PMH): We utilized past medical history, i.e. all electronic health records available before the individual's recruitment. More specifically, we utilized a boolean ICD subcategories status indicator across all ICD subcategories with a minority class share of > 0.1%. In addition to Age & Sex, this comprised $n_{features}$ = 434 ($n_{continuous}$ = 0, $n_{boolean}$ = 434) columns.

(V) Complete: We combined all aforementioned predictor matrices. This comprised $n_{features}$ = 685 ($n_{continuous}$ = 237, $n_{boolean}$ = 448) columns.

Derivation of all those data points, with their respective UKB Field IDs and descriptions, can be found in Supplementary Table 2.

### 3.4 Inclusion and exclusion criteria

Starting from the whole UKB cohort, we first included all individuals who had a respective missingness of < 20 % across all predictor matrices (thus mainly removing individuals without metabolomics coverage, leaving us with roughly *n* ≈ 240,000 participants). For each endpoint, we further excluded individuals with baseline occurrence of the respective endpoint (*vide supra* for endpoint definitions).

### 3.5 Preprocessing and Imputation

All factor variables were one-hot encoded, thus restraining all column classes to either continuous or boolean.

In the specified order, we applied three layers of missingness-criteria based exclusion: (I) exclusion of individuals with absence of an assay, i.e. 100% missingness for either clinical risk score variables, metabolite measurements or polygenic risk scores; (II) exclusion of variables with > 20% missingness across all retained individuals; (III) exclusion of retained individuals with > 20% missingness across all retained variables.

Next, we imputed missing values using a chained-equations framework with random forest predictor (package *miceRanger*). Imputation was performed according to the cross-validation (CV) approach described below separately for each of the five CV splits. Imputation models were trained on the training subset and applied to the whole dataset.

Further preprocessing was performed under the assumption of a detection threshold for biochemical measurements (i.e. variables within the clinical risk score or metabolite measurements), for which we replaced 0s with $1/10^{th}$ of the median. We performed log-scaling and centering (utilizing the train partition mean and standard deviation (SD)) for all continuous variables. For all clinical chemistry and metabolomic measurements, we assumed technical bias/artefacts for outlier measurements > 5 SD from the mean and excluded such individuals.

To avoid numerical issues during model fitting, we filtered all predictor matrices for low-variance (exclusion at a threshold of < 0.01) and perfect inter-predictor-pair correlation (to this end we calculated correlation of all input feature pairs and retained only one feature for all pairs with near-perfect correlation (R > 0.99)).

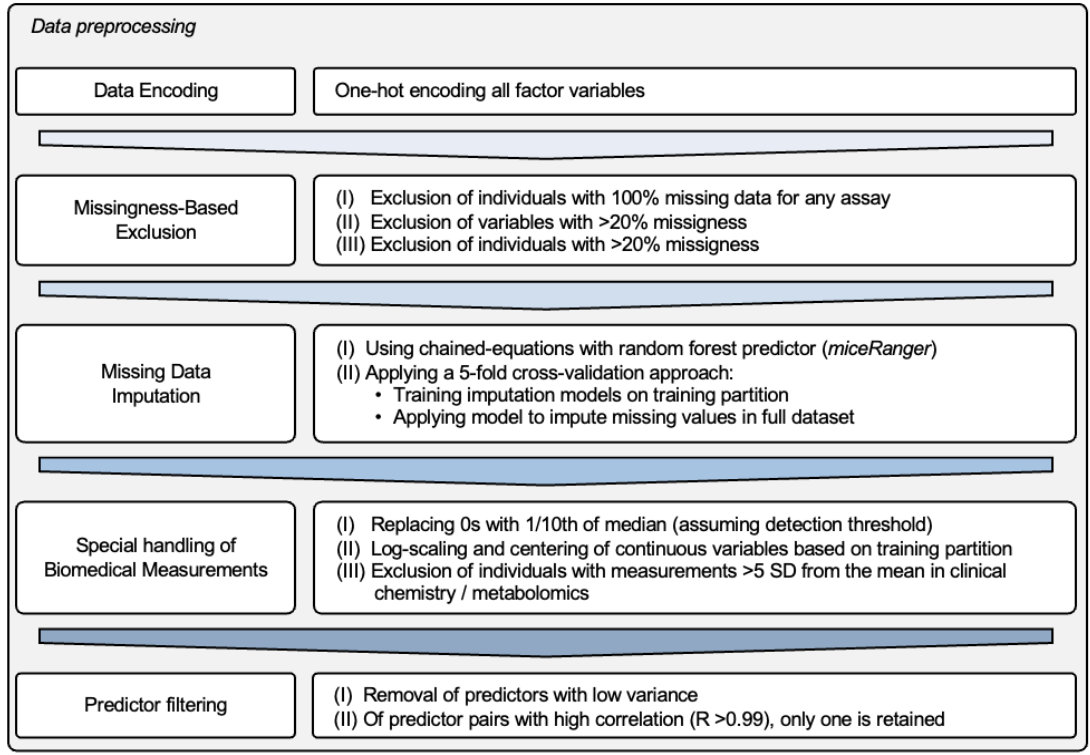

**Figure 2: Data preprocessing flowgraph.** Data preprocessing can be summarised in five successive steps: Data Encoding, Missingness-Based Exclusion, Missing Data Imputation, Special handling of Biomedical Measurements and Predictor filtering. Vertical order represents order of operations. These steps were performed within each outer Cross-Validation (CV) loop, i.e. imputation models were trained on the CV's respective training fold and subsequently applied to the full dataset. SD: standard deviation.

### 3.6 Model Fitting

#### 3.6.1 Linear methods

We used the *scikit-survival*[18] package to different variants of classical COX-PH models. First, we trained a classical COX-PH model with CoxPHSurvivalAnalysis(), applying a minimal L2 penalty (alpha = $1e^{-6}$) to avoid numerical instability in matrix inversion steps. We then fitted a Ridge-penalized COX-PH model using the same function, exploring five alpha values spaced logarithmically between $1e^{-3}$ and 1. Next, we switched to CoxnetSurvivalAnalysis() for Lasso-penalized COX-PH, setting the L1 ratio to 1 and tuning alpha over the same range. Finally, we trained Elastic Net models - again using CoxnetSurvivalAnalysis() - optimising along the same alpha grid whilst simultaneously testing ten L1 ratio values in increments of 0.1, ranging from 0.1 to 1.0. Where relevant, optimal hyperparameters were chosen via the CV strategy outlined below (see 3.7).

#### 3.6.2 Ensemble methods

<u>Random Survival Forests:</u> First, we sought to implement Random Survival Forests, as described by Ishwaran et al.[19]. We experimented with *scikit-survival*'s RandomSurvivalForest() function, however observed extremely long training times, hardly feasible for datasets of *n* > 100,000. We therefore utilized the highly optimised Ranger implementation, which was available in Python via the *skranger* package and the RangerForestSurvival() function. We optimised the models along a hyperparameter grid, growing 50, 100, or 200 trees at depths of 3, 7, or 10, respectively. The node size was restricted to a minimum of 20.

Gradient Boosting Machines: An off-the-shelf GBM approach was implemented using *scikit-survival*'s GradientBoostingSurvivalAnalysis(). Models were optimised along a hyperparameter grid comprising 50, 100, or 200 estimators and depths of 3, 7, or 10, respectively. The leaf size was restricted to a minimum of 10.

As this implementation was unfeasible for datasets with *n* > 100,000 individuals, we sought to additionally implement a survival objective function for Microsoft's highly efficient *LightGBM* framework. More specifically, we provide a gradient and hessian matrix of the negative partial log likelihood with regards to the prediction. We pre-compute risk sets, using Breslow's method for ties, for each event after sorting our dataset. Ensuring numerical stability (avoiding division by zero via incorporating a near-zero constant), we compute the gradient and hessian matrices iterating over our dataset and finally map the matrices back into the original order. Models were optimised along a hyperparameter grid comprising 50, 100, or 200 estimators and maximum leaf numbers of 7, 127, or 1023, respectively. The leaf size was restricted to a minimum of 10.

Extreme Gradient Boosting (XGB): XGB models were implemented via *xgboost*'s XGBRegressor(), utilizing the inbuilt survival task handling via objective='survival:cox' and eval_metric='cox-nloglik'. Models were optimised along a hyperparameter grid comprehensively testing 50, 100 or 200 estimators and depths of 3, 7 and 10. The minimum child weight was restricted to 10.

### 3.6.3 Deep Learning

As previous studies have shown, neural networks can be effective in predicting hazard ratios. A review by Wiegrebe et al[20] found the most common architecture are Cox-based approaches, optimising negative log-likelihood with a feed-forward neural network. In line with this and the work of Katzmann et al[21], we designed and trained a feed-forward neural network, based on repeating blocks of a fully connected layer, batch normalisation (here, we diverged from Katzmann et al.), a rectified linear unit and a dropout layer. During hyperparameter optimisation, we varied the number of hidden layers (2, 3 or 5) and number of neurons per hidden layer (16, 64, 256). Implementation details can be found in Table 1.

| Model | Package & Version | Function | Fixed Hyperparameters | Tuned Hyperparameters |
|---|---|---|---|---|
| COX-PH | scikit-survival-0.23.1 | CoxPHSurvivalAnalysis() | alpha:1e$^{-6}$ | |
| Lasso-penalized COX-PH | scikit-survival-0.23.1 | CoxnetSurvivalAnalysis() | max_iter: 100<br>l1_ratio: 1.0 | alphas: np.logspace(-3, 0, 5) |
| Ridge-penalized COX-PH | scikit-survival-0.23.1 | CoxPHSurvivalAnalysis() | | alpha: np.logspace(-3, 0, 5) |
| Elastic Net-penalized COX-PH | scikit-survival-0.23.1 | CoxnetSurvivalAnalysis() | max_iter: 100 | l1_ratio: np.linspace(0.1, 1.0, 10)<br>alphas: np.logspace(-3, 0, 5) |
| Random Forest | skranger-0.8.0 | RangerForestSurvival() | min_node_size: 20<br>seed: 42<br>oob_error: False | n_estimators: [50, 100, 200]<br>max_depth: [3, 7, 10] |
| LightGBM | lightgbm-4.5.0 | lgb.train() | objective: cox_ph_loss_lgb (custom, *vide supra*)<br>boosting_type: 'gbdt'<br>learning_rate: 0.1<br>min_data_in_leaf: 10<br>max_depth': -1, (allow unrestricted depth as tuned by num_leaves)<br>metric: 'None'<br>seed: 42<br>verbose: -1 | n_estimators: [50, 100, 200]<br>num_leaves: [7, 127, 1023] |
| XGBoost | xgboost-2.1.3 | XGBRegressor() | objective: 'survival:cox'<br>eval_metric: 'cox-nloglik'<br>tree_method: 'exact'<br>learning_rate: 0.1<br>min_child_weight: 10 | n_estimators: [50, 100, 200]<br>max_depth: [3, 7, 10] |
| Deep Learning | torchsurv-0.1.4<br>torch 2.6 | torch.nn.Module() | learning_rate: 0.001<br>lr_patience: 5<br>dropout: 0.2<br>batch_size: 50000<br>activation_function: torch.nn.BatchNorm1d<br>optimizer: torch.optim.Adam | num_layers: [2, 3, 5]<br>layer_size: [16, 64, 256] |
| GBM (sksurv) | scikit-survival-0.23.1 | GradientBoostingSurvivalAnalysis() | learning_rate: 0.1<br>min_samples_leaf: 10 | n_estimators: [50, 100, 200]<br>max_depth: [3, 7, 10] |

**Table 1: Model Implementation details.** Alongside package versions and the model training function we used, we also provide both fixed and tuned model hyperparameters. We do not explicitly provide default hyperparameters if they were left unchanged (e.g. n_iter: 100 for *scikit-survival*'s CoxPHSurvivalAnalysis()). COX_PH: COX proportional hazards, GBM: gradient boosting machine.

3.7 Cross-validation strategy for hyperparameter tuning and performance evaluation

To facilitate hyper-parameter tuning alongside model evaluation, we utilised a 5-fold, nested CV approach. The outer 5-fold CV loop consisted of an 80% train and a 20% test split, with imputation model (vide supra) and prediction model (vide supra) training being performed on the training subset ($n \approx 200,000$ individuals), whereas performance metrics were evaluated within the test subset ($n \approx 50,000$ individuals). An inner, 5-fold CV loop was used to divide the outer loops training subset into 80% training and 20% testing splits, here utilised to evaluate model hyperparameters: Along a hyperparameter grid (see below), one model per unique hyperparameter combination was fitted on each training partition and evaluated for Harrel's C on the test individuals. We chose the best performing hyperparameter combination for its best average performance across the inner CV loop.

For the penalized COX-PH models, we tuned the penalty coefficients along a broad range of 1e$^{-3}$ to 1 with logarithmic spacing. L1/L2 ratios were tuned in 10 linear steps between 0.1 and 1.

For all non-linear models, we chose the hyperparameter space in a manner to restrain complexity to comparable sizes. A subset of hyperparameter combinations with minimal, maximal and intermediate complexity are shown in Table 2. In addition to the displayed hyperparameter combinations, all possible hyperparameter combinations (as outlined in 2.5.; vide supra) were tested in a 5-fold, internal-to-training CV approach.

| Model | Hyperparameter Configurations | Approximated Complexity |
|---|---|---|
| Random Forest | n_estimators = 50, max_depth = 3 | 50×7=350 |
| | n_estimators = 100, max_depth = 7 | 100×127=12,700 |
| | n_estimators = 200, max_depth = 10 | 200×1023=204,600 |
| LightGBM | n_estimators = 50, num_leaves = 7 | 50×7=350 |
| | n_estimators = 100, num_leaves = 127 | 100×127=12,700 |
| | n_estimators = 200, num_leaves = 1023 | 200×1023=204,600 |
| XGBoost | n_estimators = 50, max_depth = 3 | 50×7=350 |
| | n_estimators = 100, max_depth = 7 | 100×127=12,700 |
| | n_estimators = 200, max_depth = 10 | 200×1023=204,600 |
| Deep Learning | num_layers = 2, layer_size = 16 | $2 \times 16^2$=512 |
| | num_layers = 3, layer_size = 64 | $3 \times 64^2$=12,288 |
| | num_layers = 5, layer_size = 256 | $5 \times 256^2$=327,680 |

**Table 2: Subset of utilised hyperparameter settings for non-linear models.** Complexity approximates and their derivation (e.g. n_estimators × $2^{max\_depth}$ for tree-based models) are showcased for the model's respective minimal, maximal and intermediate complexity hyperparameter settings.

Discrimination, rather than calibration, is of primary importance to guide preventive efforts, where artificial or age-group-dependent risk thresholds essentially aim at identifying at high-risk collective. Moreover, specifically in the context of primary prevention efforts at population scale, efficiently allocating a limited number of resources (e.g. for follow-up screenings) to the individuals at highest relative risk is crucial. At the scale benchmarked here, even in settings where we want to weigh the treatment benefits vs. the risk of adverse effects, the average absolute risk of e.g. the highest risk decile can be assumed to be relatively constant. Finally, risk scores possessing good discrimination can be recalibrated, whereas a well-calibrated score that doesn't discriminate well is essentially useless to guide clinical decision making. In contrast to other authors[10], we therefore intentionally focused our study on various metrics of discriminative performance. Considering its widespread usage[22] and interpretability, we assessed Harrel's C-index as the primary performance metric alongside a range of metrices (Uno's C-index alongside Sensitivity, Specificity, False Negative and Positive Ratios, Hazard ratios and Restricted Mean Survival Times for the top 10% and 20%, respectively) assessing discrimination on the test split. We are aware that Harrel's C is occasionally criticised for its bias in settings of moderate-high censoring[23,24] (which is not the case in our instance), and therefore also present Uno's C index as a more stable alternative (in Supplementary Table 3). Moreover, overfitting bias was calculated as ΔC between outer CV loop training and testing splits. The range of hyperparameter values for each of the methods were selected to cover a wide range but also to assure a fair comparison across methods (identical number of estimators grid for the methods where number of estimators is a critical parameter).

### 3.8 Significance, software and code availability

Models were compared using paired t-tests and false discovery rate (FDR)-adjusted using the Benjamini-Hochberg procedure. Significance was defined as FDR-adjusted $p < 0.05$. All experiments were conducted on UKB's Research Analysis Platform, an Amazon Web Services (AWS) - empowered platform, hosted by DNAnexus®. Different parts of the code were performed in R 4.4.0 and Python 3.9.16 (all model fitting was performed in Python). Large language models were used to assist with coding steps and proof-reading. All code will be made publicly available via GitHub upon publication.

### 3.9 Data availability

The UKB dataset is publicly available to approved researchers at https://www.ukbiobank.ac.uk/enable-your-research. Definitions for endpoints and predictor matrices can be found in Supplementary Tables 1 and 2, detailed performance metrices can be found in Supplementary Table 3.

4. Results

4.1 Performance across endpoints and input features

We first ran a model training pipeline in a nested 5-fold CV approach, in which the outer CV loop was used to perform model fitting and evaluation across endpoints and input feature combinations, and an internal-to-training CV loop was used to optimise model hyperparameters. For a variety of linear and non-linear models, this was conducted on UKB's research analysis platform using identical compute clusters.

First, we show absolute Harrel's C-Index (Figure 3A, left panel) and ΔC relative to the COX-PH model (Figure 3A, right panel), across endpoints for the distinct input feature combination "Clinical Risk". Results were least variable for the cardiovascular disease (CVD) endpoint, which had relatively large event numbers as compared to the simultaneously assessed breast cancer (BC) and Alzheimer's disease (AD) endpoints. Here, the DL model performed best (mean Harrel's C 0.721 [95% confidence interval [CI] 0.718-0.723], mean ΔC of 0.002 [95% CI 0.001-0.002]) across CV-splits, indicating a slight improvement over the benchmark COX-PH model (q = 0.018). In contrast, the XGB model exhibited lowest C-Indices (mean Harrel's C 0.564 [95% CI 0.560-0.569], mean ΔC –0.155 [95% CI –0.159-–0.151]) across CV-splits. Notably, results varied more for BC (overall relatively low C-indices; highest Harrel's C for Elastic Net (0.563 [95% CI 0.553-0.573]), COX-PH (0.563 [95% CI 0.557-0.570]), and Ridge (0.563 [95% CI 0.557-0.569]) models) and AD (overall relatively high C-indices; highest Harrel's C for Ridge (0.838 [95% CI 0.834-0.842]), COX-PH (0.838 [95% CI 0.834-0.842]) and Elastic Net (0.837 [95% CI 0.833-0.841]) models) endpoints. In line with the CVD endpoint, XGB consistently showed the poorest discriminative performance (BC: Harrel's C 0.514 [95% CI 0.509-0.519]; AD: Harrel's C 0.576 [95% CI 0.554-0.598]).

A ranking of all ML models across all endpoints and input features is depicted in Figure 3B. Here, we ranked the models based on Harrel's C within each endpoint, input feature set and CV split combination.

When trained on the most extensive input feature set "Complete" (including age, sex PRS, metabolomics, PMH and clinical risk score predictors), Ridge and Elastic Net achieved consistently high rankings. Out of 15 available first-place rankings, Ridge and Elastic Net both ranked first in 7 instances. Lasso is the only other ML model that ranked first, achieving this once (for BC). Both Lasso and COX-PH models also demonstrated consistently strong discriminative capacities, scoring 6 and 4 second-place rankings, respectively. DL, GBM and Random Forest performed similarly, ranking mainly between third and seventh while XGB consistently ranked worst in this input feature combination.

A similar trend can be seen for other input feature combinations. Notably, across most of them, Ridge, COX-PH, Elastic Net and Lasso perform amongst the top ranks. Random Forest, DL and GBM show varying placements across all ranks, while XGB consistently ranks last. Interestingly, training on the "Age & Sex" and "Clinical Risk" feature combination, GBM and DL also exhibit strong performance, scoring multiple first ranks each.

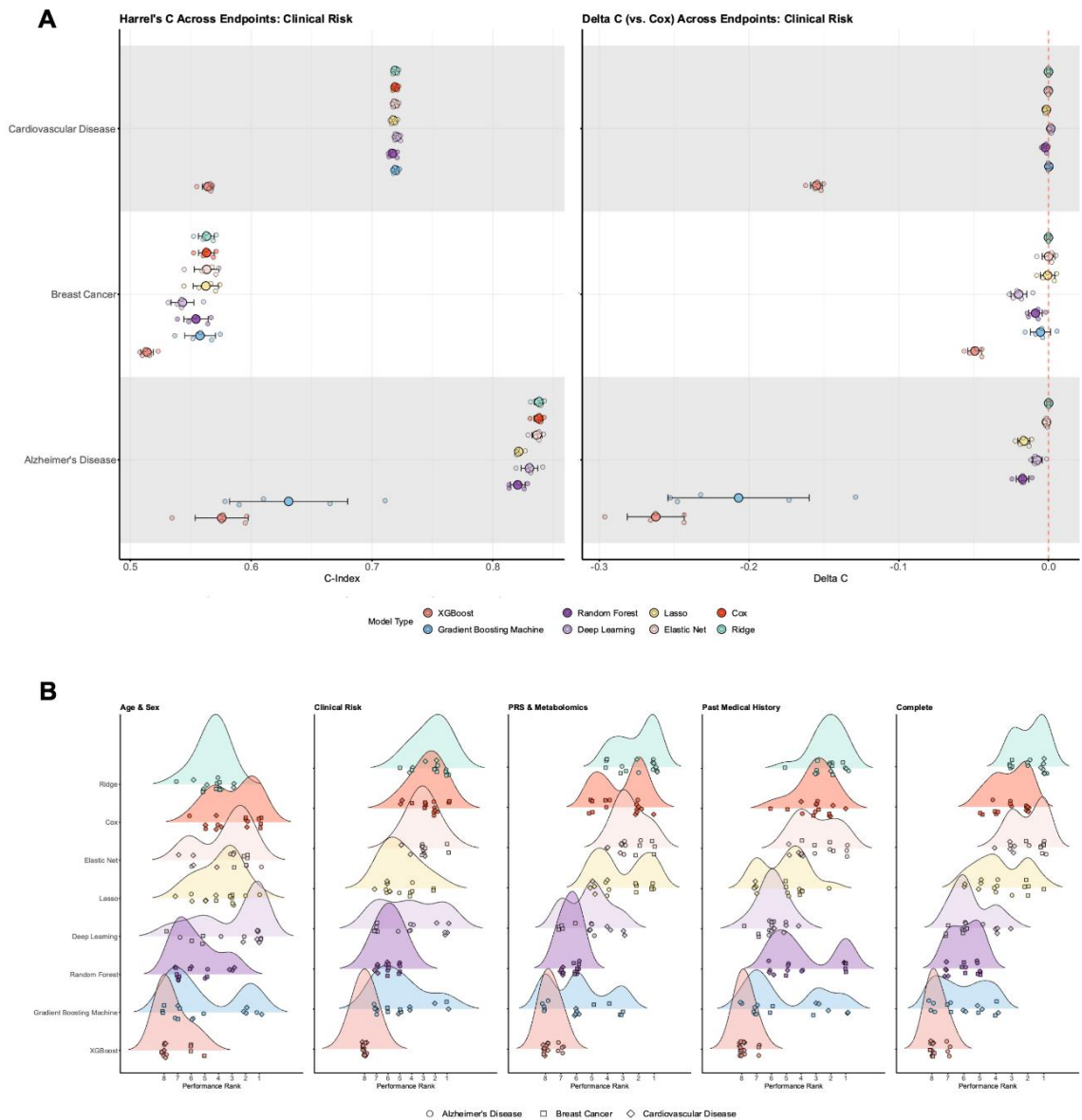

**Figure 3: Discrimination across predictor matrices and endpoints:** Absolute Harrel's C (Panel A, left side) and ΔC (Panel A, right side; relative to the COX Proportional Hazards (COX-PH) model) across endpoints for the input feature combination "Clinical Risk". Mean Harrel's C and ΔC across Cross-Validation (CV) splits are shown as large dots, small dots represent individual CV split data points, error bars represent 95% confidence intervals. In Panel B, models are ranked based on Harrel's C within each endpoint, input features, CV split combination; the subplots show ranked Harrel's C within one input feature combination (rank 1 indicating highest Harrel's C). PRS: polygenic risk scores.

### 4.2 Computational Resources Requirements

We evaluated computing time requirements by measuring the final model training time using the previously optimised hyperparameter set. Whilst the absolute training times may seem low, it is usually vital to perform CV and hyperparameter tuning (as done here), which substantially increases the number of total model training iterations that must be performed. In this context, we included two horizontal dashed lines in Figure 4A: a red line representing a cost of £0.1 per model fit and an orange line representing a cost of £0.01 per model fit.

In the upper left subplot, the COX-PH model's computing time is depicted. When training on the "Age & Sex" input matrix, computing times ranged from 3.576 s [95% CI 3.525-3.627] to 10.953 s [95% CI 10.068-11.837]. Computing times for "Clinical Risk" were slightly higher (6.341 s [95% CI 4.648-8.034] to 13.175 s [95% CI 12.303-14.047]), followed by the "Past Medical History" fitting times that ranged from 6.443 s [95% CI 6.409-6.477] to 44.508 s [95% CI 43.036-45.981]. Fitting times for "PRS & Metabolomics" ranged from 18.082 s [95% CI 17.999-18.166] to 52.013 s [95% CI 49.025-55.002]. For the most extensive input feature combination "Complete", fitting times ranged from 82.793 s [95% CI 64.924-100.662] to 465.976 s [95% CI 452.912-479.039]. Very similar computing times patterns were observed for Ridge and XGB.

The DL model exhibits relatively consistent fitting times across all input feature combinations (ranging from 47.444 s [95% CI 32.620-62.268] for "Clinical Risk"/BC to 162.463 s [95% CI 132.331-192.594] for "Clinical Risk"/CVD). Similarly, GBM training time was hardly dependent on input feature properties (ranging from 6.826 s [95% CI 6.798-6.855] for "Age & Sex"/BC to 45.754 s [95% CI 35.855-55.654] for "Clinical Risk"/CVD). Particularly short fitting times were observed for Elastic Net and Lasso models, ranging from 0.020 s [95% CI 0.020-0.020] for "Age and Sex"/BC/Lasso models to 9.907 s [95% CI 8.494-11.321] for the "Complete"/CVD/Elastic Net model.

To further synthesize our results, we performed per-predictor matrix ranking of average model fitting time across endpoints (Figure 4B). Rank 1 corresponds to the shortest fitting time. In this comparison, Lasso is quickest for "Clinical Risk", "PRS & Metabolomics", "Past Medical History" and "Complete" while Elastic net is fastest for "Age & Sex" and ranks second for all other input feature combinations. Other model types are shown to trade ranks across input feature combinations. Models displaying consistent fitting times even for larger matrices, such as DL, start being competitively fast compared to linear models for larger predictor matrices (rank 5 for the "Complete" input feature combination).

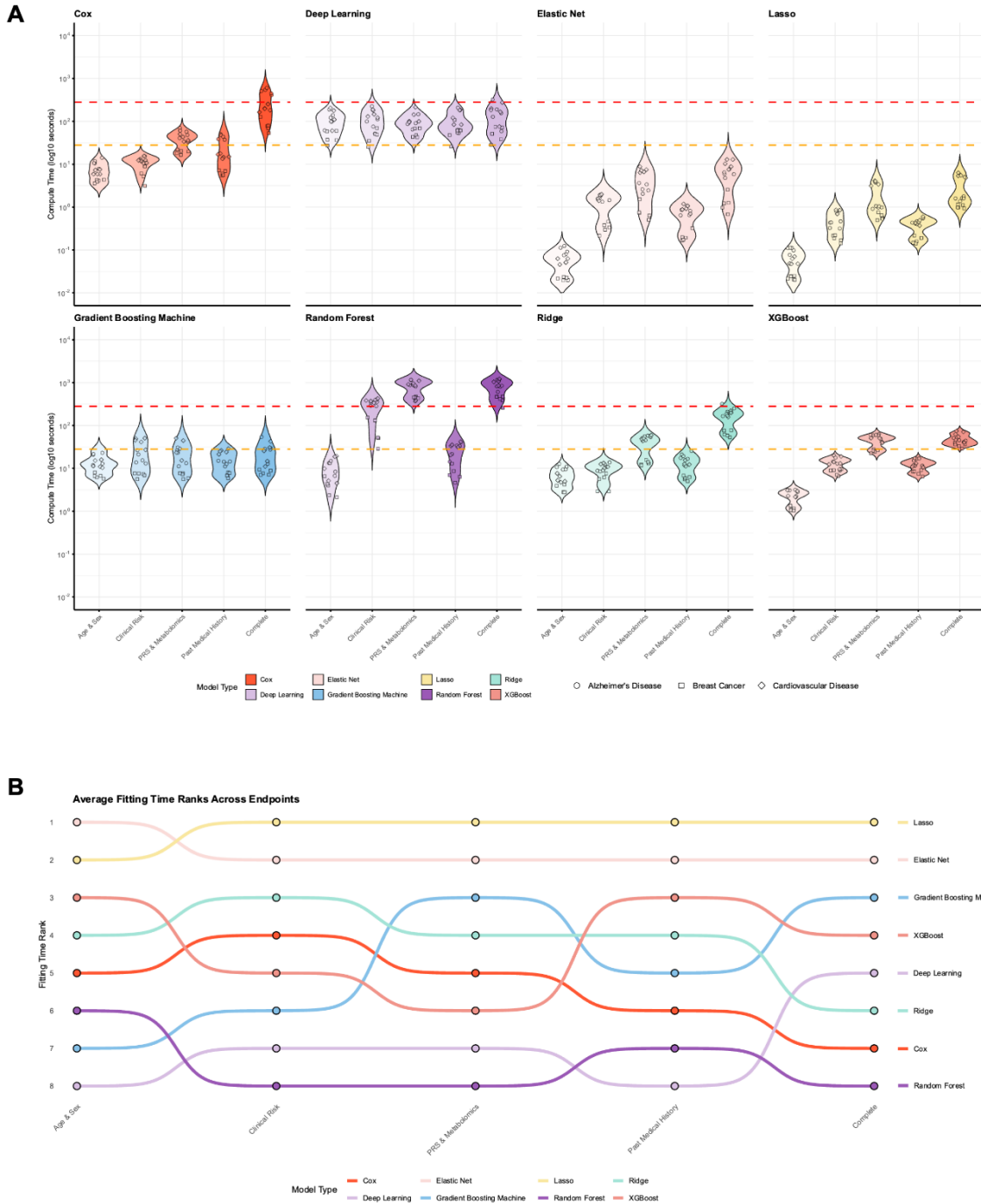

**Figure 4: Compute time across predictor matrices and endpoints:** Model training time on identical Amazon Web Services instances across endpoints (denoted by marker shape), input feature combination (horizontal axis) and model type (subplots) (A). We marked two thresholds with horizontal dashed lines: a red line representing a cost of £0.1 per model fit and an orange line representing a cost of £0.01 per model fit. Panel B depicts a ranking of model types according to their fitting time (averaged across endpoints and Cross-Validation splits). Rank 1 corresponds to the shortest fitting times. PRS: polygenic risk scores.

### 4.3 Overall Model Ranking and additional metrics

To facilitate choosing the optimal model type according to the available input features, the desired performance metric or available compute, we condensed all our findings in ranked heatmaps in Figure 5A. Similarly to the results in Figure 3B, Linear models (Ridge, COX-PH, Elastic Net, Lasso) exhibit high average Harrel's C results across endpoints and input features. XGB is outperformed in all combinations and GBM, Random Forest and DL exhibit varied performance with top rankings in specific instances.

In Figure 5B, we depict the ranking of ML models per endpoints and a range of discrimination metrics by averaging the results over CV-splits and then ranking within each metric, endpoint and model combination. Specifically, we show Harrel's and Uno's C, as well as a set of metrics for the top two risk deciles: confusion matrix metrics (Sensitivity, Specificity, False Positive Rate, False Negative Rate), the significance in a logrank test, ΔRestricted Mean Survival Time as well as Overfitting Bias (ΔHarrel's C; light gray background) and Fitting Time (gray background).

Generally, the results are very much aligned with our observations for Harrel's C. Linear models (Ridge, COX-PH, Elastic Net) perform well across all metrics (and endpoints), DL and GBM perform very well or even best for the CVD endpoint; Random Forest and XGB exhibit the worst performance in comparison to all other model types. Additional plots and metrics for the top decile can be found in Supplementary Figures 2 and 3, Supplementary Table 3, showing highly similar results.

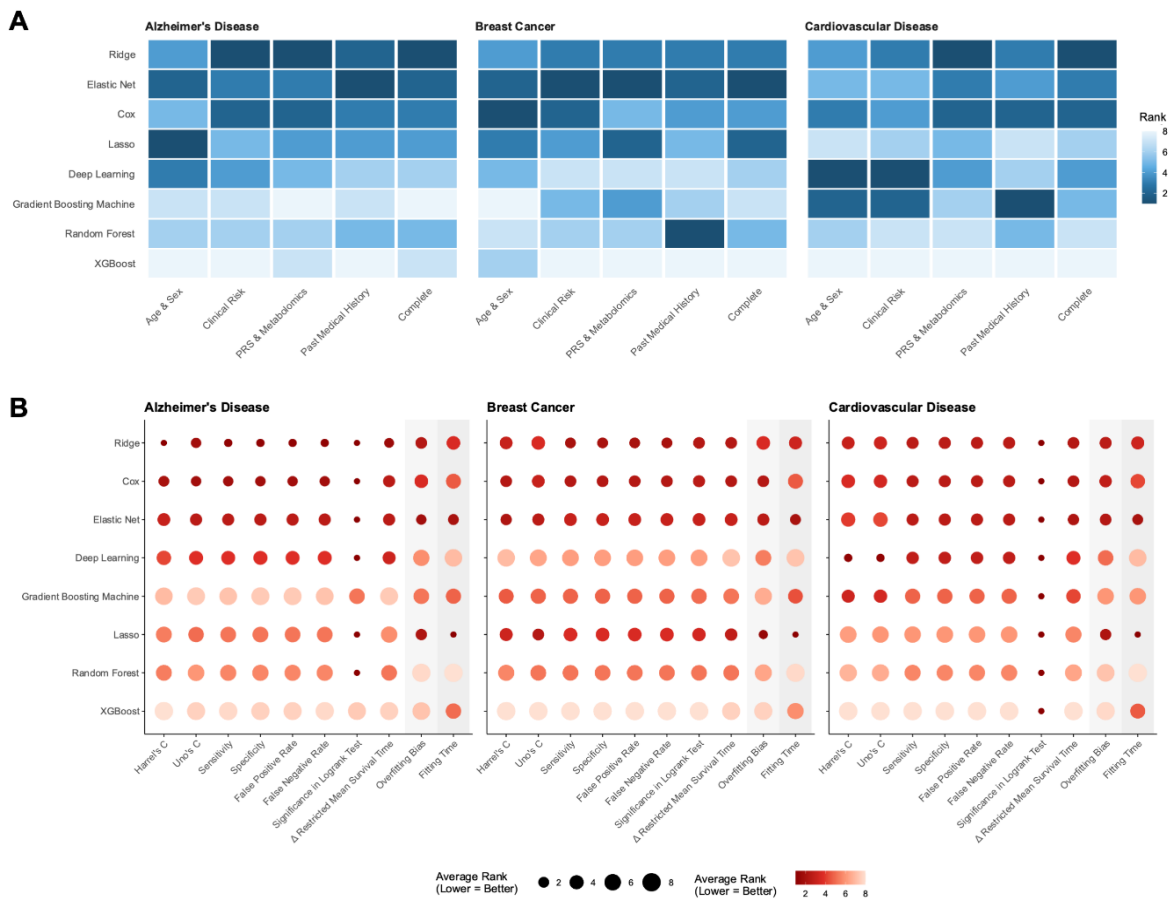

**Figure 5: Condensed overview of benchmarking results:** Average Harrel's C ranks are displayed in per-endpoint subplots in Panel A; darker colours represent lower and better rank. Panel B showcases

ranking of machine learning models for various metrics, in per-endpoint subplots for the input feature combination "Clinical Risk"; darker colour and smaller dots represent lower and better rank. PRS: polygenic risk scores.

### 4.4 Scaling with sample size

To further investigate the link between sample size and model discrimination as well as computing time, we performed the same experiment with random subsets of our cohort, thus incrementally restricting the sample sizes to $n$ = 5,000, 10,000, 20,000, 50,000, respectively. In Figure 6A, we show Harrel's C (averaged across CV splits) for the predictor combination "Complete" (results for all predictor combinations can be found in Supplementary Figure 4). Result variability generally declined with increasing number of samples and observations: the consequently observed variability of the AD endpoint was higher compared to the other endpoints and sample size $n$ = 5,000 resulted in the highest variability across endpoints. The discrimination ranking of models is comparable to the ranking for the full dataset, with linear models generally performing well and XGB performing worst in most cases. Notably, DL and *LightGBM* models exhibit increasingly competitive performance with increasing sample size. This trend is especially apparent for the CVD endpoint (high event counts), where DL initially scores the second lowest Harrel's C for $n$ = 5,000 samples, but then minimizes the discriminative gap with increasing sample size.

The *scikit-survival* GBM implementation is included in this analysis alongside the *LightGBM* implementation. In Panel B, we show the fitting time of all model types for each sample size and endpoint combination, however we again show only the "Complete" predictor combination, with all further results available in Supplementary Figure 5. Training time increases with increasing sample size in line with expectations (logarithmic y-axis, a linear increase in the figure space corresponds to an exponential increase of fitting time with increasing sample size). Here, the required computational resources for the scikit-survival GBM implementation were drastically higher than for all other models already for $n$ = 50,000 samples, thus necessitating utilisation of different strategies for the full dataset (*LightGBM* shown to be approximately $10^2$ times faster than *sksurv)*.

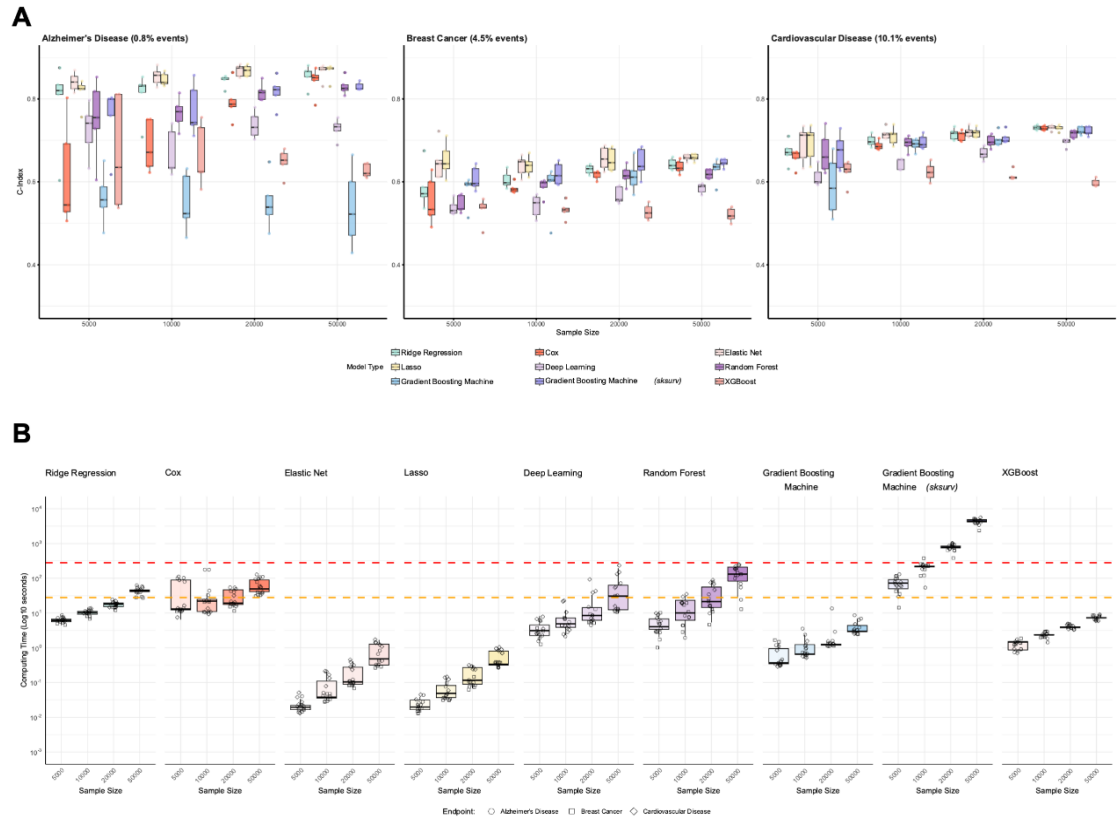

**Figure 6: Results from smaller sample sizes:** Panel A showcases absolute Harrel's C-Index per endpoint and model type over sample size for the predictor matrix "Complete". Boxplots show the median as horizontal line, span the inter quartile range and have whiskers ranging from the 5[th] to the 95[th] percentile. Panel B depicts the computing time per model fit (averaged across Cross-Validation splits), data point shapes differentiate endpoints. Two thresholds were marked with horizontal dashed lines: a red line representing a cost of £0.1 per model fit and an orange line representing a cost of £0.01 per model fit.

5. Discussion

Prevention is vital to ensure healthy ageing and to alleviate the ever-growing burden imposed upon healthcare systems by chronic disease and an increasingly multimorbid, ageing society. Accurate risk stratification is vital to guide the effective implementation of primary and secondary prevention efforts. With the availability of deeply phenotyped, population-scale biobanks, and specifically also the cost-effective integration of different high-throughput -omics technologies, researchers are now provided with unprecedented opportunities for risk modelling. Moreover, researchers are able to leverage a rich toolkit of open-source and available ML frameworks, many being equipped with off-the-shelf survival functionalities, with distinct advantages in specific scenarios.

To date, several studies benchmarked different risk modelling strategies on highly heterogenous, publicly available datasets. For example, Herrmann et al. benchmarked various survival analysis tools on a variety of high-dimensional, small-scale cancer datasets, showing very robust performance of standard COX-PH models[10]. Similar results were obtained by Zhang et al, who also analysed small-sized datasets (or random, small subsets of larger studies), explaining the relatively high variability of the results. Unfortunately, without any hyperparameter tuning being performed in this study, the default settings used might artificially favour certain architectures[11]. Most recently, Burk et al. presented a very elegant study assessing different survival toolkits in a neutral and comprehensive fashion. Whilst this piece of work undoubtedly provides very important insights, again highlighting the very robust performance of classical linear models, the study was conducted on datasets of almost exclusively small size (mostly $n$ < 1,000 individuals, one dataset with $n$ = 50,000 individuals)[9]. As discussed below, we showcase how larger sample sizes significantly influence discriminative performance across a range of endpoints and predictor matrices. Available benchmarking efforts are in this context not necessarily reflective of the resources that many researchers nowadays wish to leverage, thus underscoring the need for thorough benchmarking at the scale presented in this study.

In line with what has previously been reported[9,10] (vide supra), our results suggest that linear models, specifically penalized COX-PH models, generally serve as a scalable, interpretable, and robust method, also for large-scale survival tasks. This is specifically valid in cases where increasing the dataset's dimensionality favours significant overfitting bias for a range of non-linear ML algorithms. Of note, there were exceptions in specific niche scenarios. For example, utilizing standard deep neural network architectures, we observed marginal discriminative performance increases specifically for datasets with large event counts and low dimensionality, such as e.g. shown for CVD prediction using a standard clinical risk score panel. Such inter-endpoint-/-predictor-matrix variability highlights the importance of benchmarking model performance comprehensively across a range of endpoints and heterogenous predictor matrices.

Simultaneously, we report the computational requirements on moderate-sized AWS instances (and thus also the financial burden). Whilst these are almost negligible for penalized linear models, extensive hyperparameter grid searchers in a CV approach, as performed here, can produce significant cost for more complex model types. Even algorithms with highly effective backend implementations, such as e.g. the *Ranger* RF package, might struggle with large predictor matrices (particularly for continuous variables) comprising $n \approx 200{,}000$ individuals.

Moreover, via stepwise restriction of sample size ($n_{individuals}$), we evaluated how well different model types, hyperparameter combinations and predictor matrices discriminate risk across endpoints. Our results show that penalized COX-PH models perform robustly even in settings of $n_{individuals} = 5{,}000$, whilst some frameworks, including a hardly penalized COX-PH model and several non-linear approaches, struggle with overfitting, which is expectedly aggravated for endpoints with low event counts. We further show how certain implementations, which might perform very well on smaller datasets, are rendered impracticable solely due to the computational burden (e.g. the *scikit-survival* GBM implementation). To address this, we respond to repeated community requests and provide a survival functionality for *LightGBM*, enabling researchers to perform fast and robust gradient boosting at scale. The reported differences in how well models scale with the number of available datapoints highlight the necessity for a given study to adjust to benchmarking results from datasets of similar size and properties.

Importantly, the work presented here constitutes an immense computational effort, thus also underscoring how exhaustive per-study benchmarking of different survival task frameworks (as repeatedly suggested for smaller datasets) becomes unfeasible for most researchers. Our work therefore fills a critical gap, now allowing researchers working on similar datasets to accurately choose methodology with simultaneous consideration of discriminative performance and computational/financial feasibility. Instead of further high-level comparative model-assessment, we envision that this should free up time and resources which can be dedicated to find further marginal performance improvements (for example via deploying more complex model architectures, performing more extensive hyperparameter tuning, ensemble and wrapper dimensionality techniques and trialling different preprocessing steps, etc.).

Our study has specific strengths and limitations: (I) To our knowledge, this study represents the largest survival task benchmarking effort to date and thus allows exclusive insights into survival task model performance at scale. However, as datasets continue to grow[25], follow-up studies will become necessary. (II) Our model configurations and hyperparameter grids were chosen with intent, specifically to cover a wide range of relatively simple to complicated models, with comparable complexity across model types (vide supra, Table 2). Considering the computational requirements, it wasn't possible or envisioned to trial all possible model configurations. Specifically for the benchmarked DNNs, which showed superior discrimination in certain settings, it is possible that further architecture modifications (e.g. different activation functions, more sophisticated early stopping criteria, weight averaging, etc.) might lead to further performance increases. (III) Similar arguments apply for model choice, where we tried to select the most commonly used frameworks. However, we are aware that the list of benchmarked models is non-exhaustive and with unlimited resources, it might have been interesting to further expand. (IV) We chose a comprehensive CV strategy ensuring repeated model evaluation on unseen data and, amongst our CV splits generally observed very consistent results. Despite the fact the external validation was out of scope for the present manuscript, any practical future application of the fitted models would require their validation in external cohorts. (V) We benchmarked models across various discriminative performance indicators, emphasizing discrimination as paramount in biomedical contexts where risk stratification into high- and low-risk groups is critical for clinical decision making (as discussed above).

In conclusion, we performed extensive benchmarking of different survival task toolkits within UKB, one of the largest prospective cohort studies to date. Ultimately, we hope this effort will empower researchers, clinicians, and data scientists to more effectively leverage such datasets for prevention and risk prediction, thereby contributing to a future in which healthcare decisions are grounded in robust, data-driven insights. The two key take-homes of this work are: (I) The performance of specific frameworks is largely dependent on the number of individuals, the endpoint frequency, and the predictor matrix dimensionality and properties. All these factors should be jointly considered (and benchmarked alongside each other, as done here), when choosing an appropriate method for a specific research question. Within this manuscript, we provide explicit instructions for selecting the best performing model in each scenario as well as a fully implemented scalable and documented implementation to support application of each selected method for these scenarios. (II) Penalized COX-PH models frequently outperform more sophisticated methodology. We therefore suggest that these represent a very robust starting point for most research questions and, specifically also considering their ease of use, interpretability and scalability, should be reported as benchmarks in publications proposing novel risk scores using more complex non-linear ML models.

## 6. Acknowledgments

The dataset was accessed under application ID 98729. We want to thank all UKB-enrolled participants for their participation.

## 7. Funding

RRO is supported by a British Heart Foundation (BHF) PhD studentship (FS/19/58/34895), BHF Pump Prime funding (RE23446) and both UKB's "getting started" and "enhanced" credit programme. RAS receives funding from the National Institute for Health and Care Research (NIHR) which supports his Academic Clinical Fellow post. AZ is supported by a BHF project grant (PG/24/11770). KT is supported by a BHF project grant (PG/20/10387). AMS is supported by the BHF (CH/1999001/11735, RG/20/3/34823, RE/18/2/34213).

## 8. Conflicts of interest

AMS serves as an adviser to Forcefield Therapeutics. All other authors have nothing to disclose.

# 9. Supplementary Data

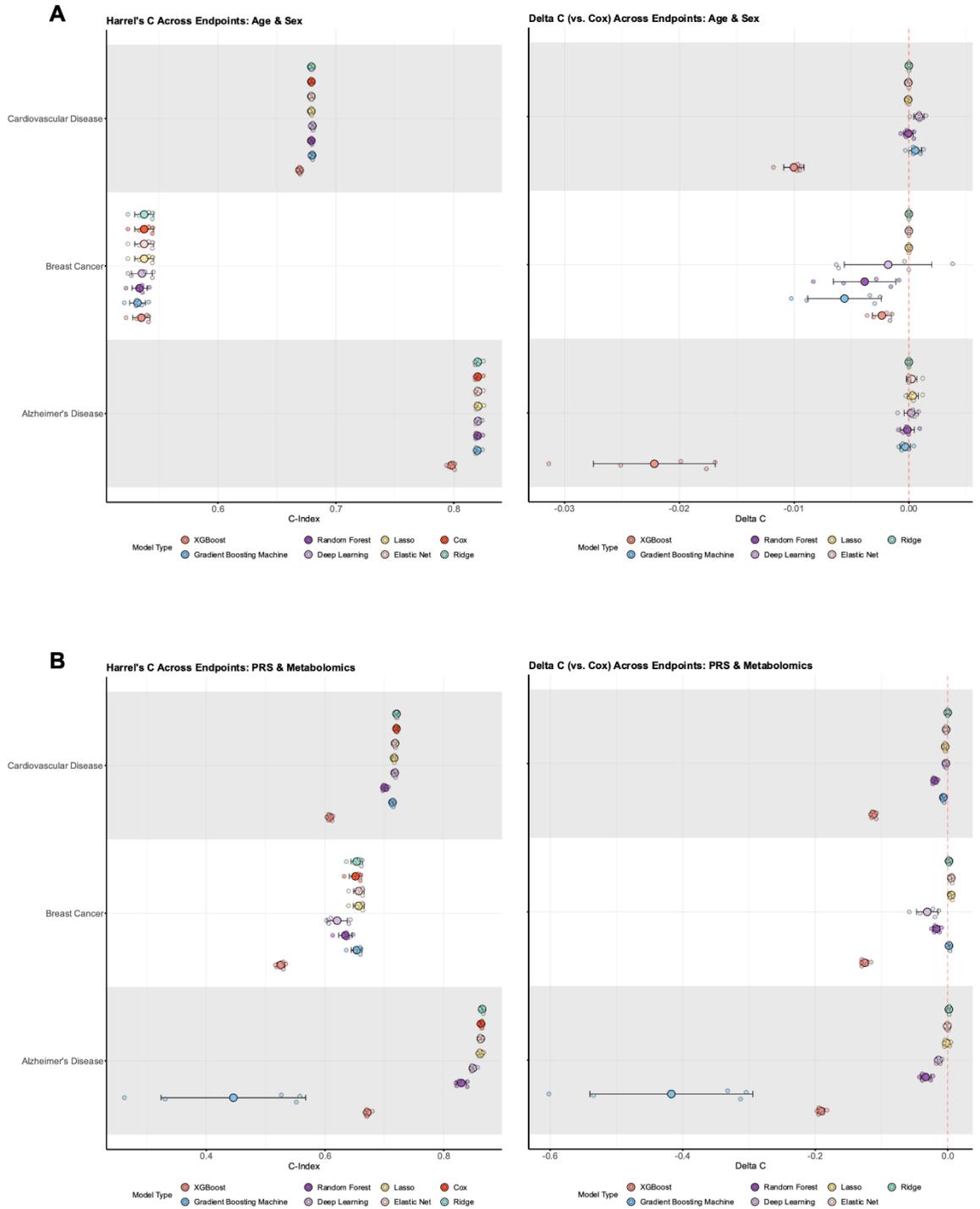

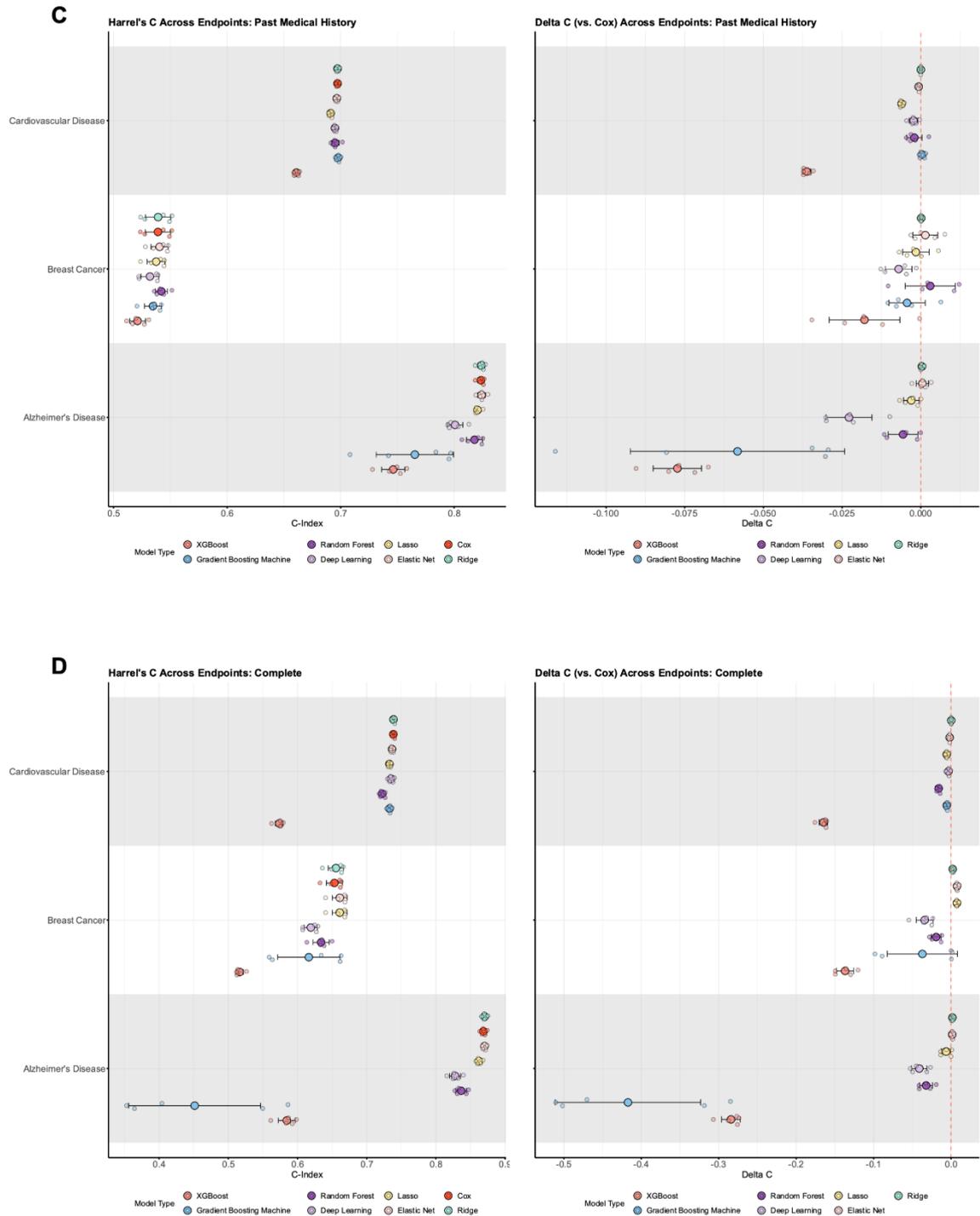

**Supplementary Figure 1: Discrimination across predictor matrices and endpoints:** Absolute Harrel's C (Panels A-D, left side) and ΔC (Panels A-D, right side; relative to the COX Proportional Hazards (COX-PH) model) across endpoints for the input feature combination "Age & Sex", "PRS & Metabolomics", "Past Medical History" and "Complete". Mean Harrel's C and ΔC across Cross-Validation (CV) splits are shown as large dots, small dots represent individual CV split data points, error bars represent 95% confidence intervals. PRS: polygenic risk scores.

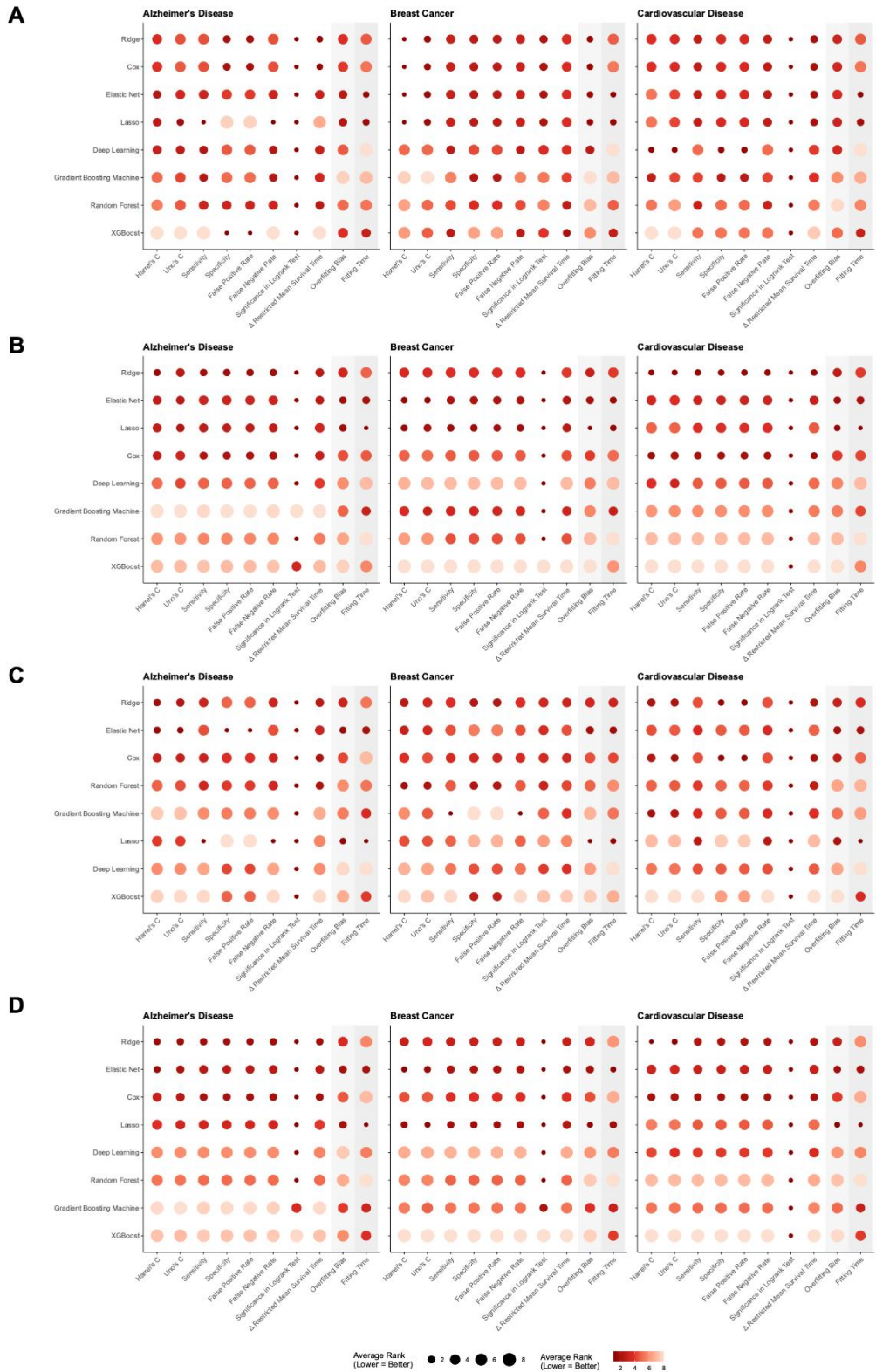

**Supplementary Figure 2: Multi-metric ranked overview of benchmarking results (high-risk cutoff: top 10%):** Panel A-E showcases ranking of machine learning models for various metrices (where applicable, calculated for the highest risk decile) in per-endpoint subplots for the input feature combinations "Age & Sex" (Panel A), "Clinical Risk" (Panel B), "Polygenic Risk Scores (PRS) & Metabolomics" (Panel C), "Past Medical History" (Panel D) and "Complete" (Panel E); darker colour and smaller dots represent lower and better rank.

**Supplementary Figure 3: Multi-metric ranked overview of benchmarking results (high-risk cutoff: top 20%):** Panel A-E showcases ranking of machine learning models for various metrices (where applicable, calculated for the highest two risk deciles) in per-endpoint subplots for the input feature combinations "Age & Sex" (Panel A), "Polygenic Risk Scores (PRS) & Metabolomics" (Panel B), "Past Medical History" (Panel C) and "Complete" (Panel D); darker colour and smaller dots represent lower and better rank.

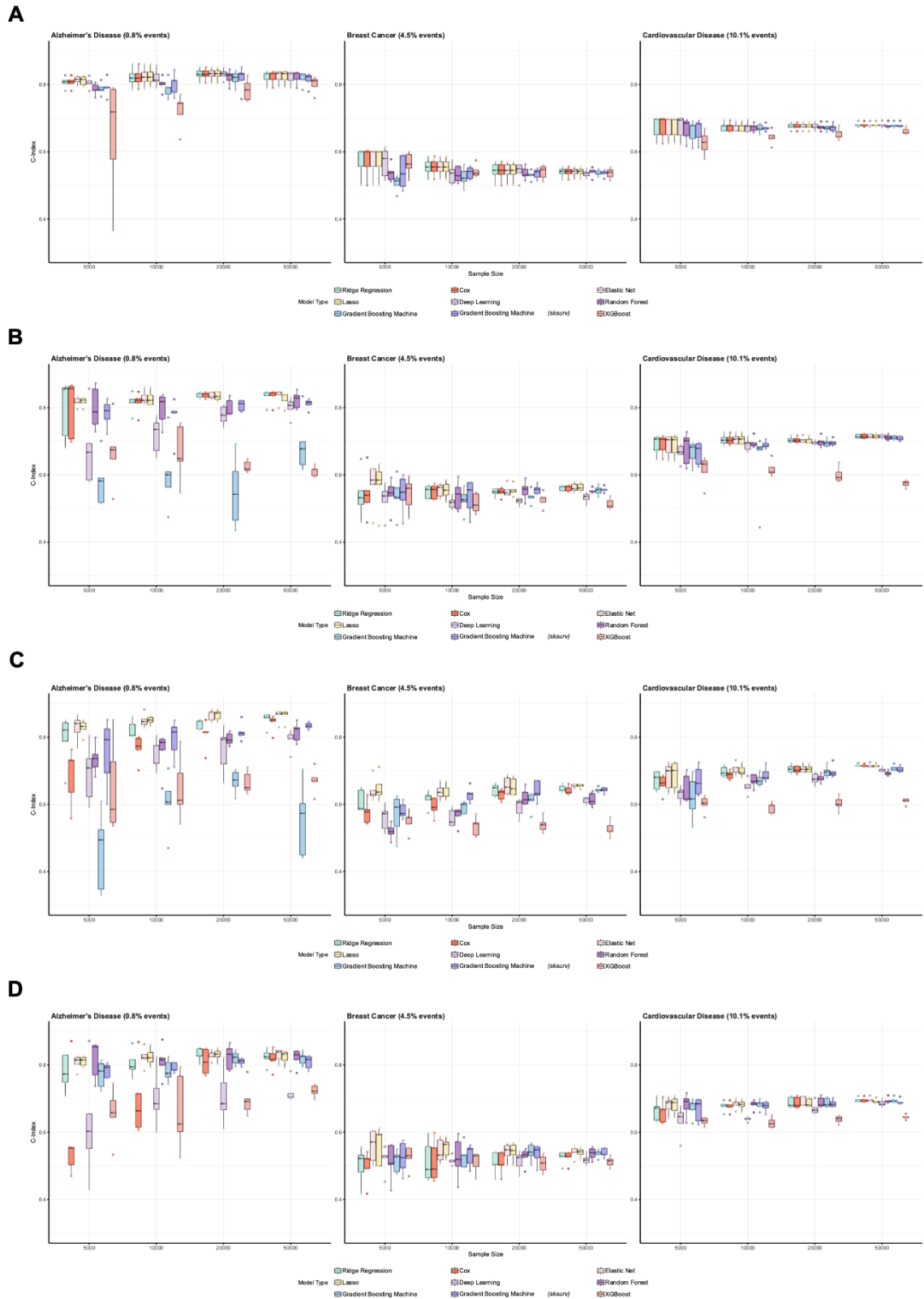

**Supplementary Figure 4: Discrimination with smaller sample sizes:** Panels A-D showcase absolute Harrel's C-Index per endpoint and model type over sample size for the predictor matrices "Age & Sex" (A), "Clinical Risk" (B), "Polygenic Risk Scores (PRS) & Metabolomics" (C) and "Past Medical History" (D). Boxplots show the median as horizontal line, span the inter quartile range and have whiskers ranging from the 5th to the 95th percentile.

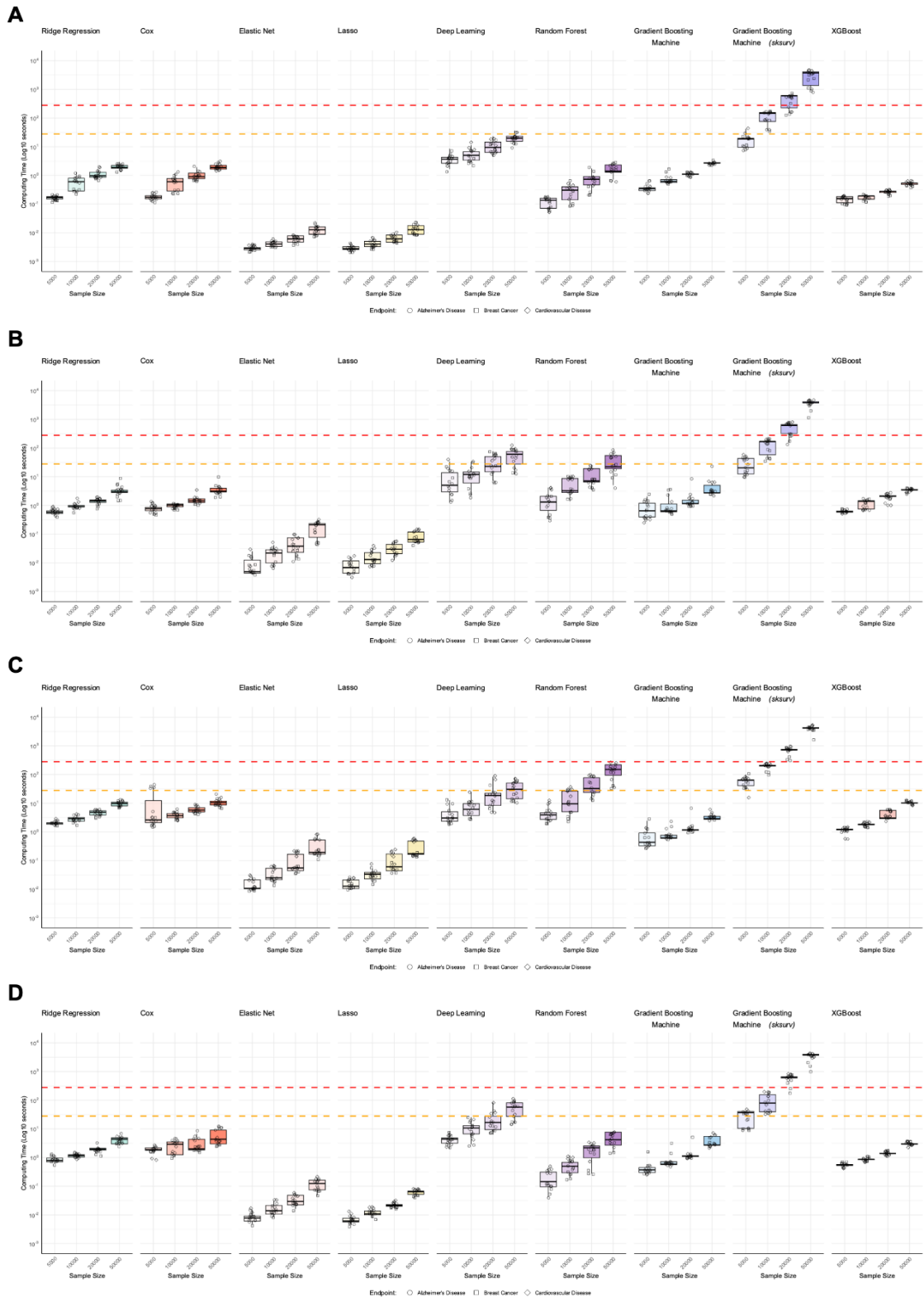

**Supplementary Figure 5: Results from smaller sample sizes:** Panel A-D depict the computing time per model fit (averaged across CV splits), across predictor matrices "Age & Sex" (A), "Clinical Risk" (B), "Polygenic Risk Scores (PRS) & Metabolomics" (C) and "Past Medical History" (D). Data point shapes differentiate endpoints. Two thresholds were marked with horizontal dashed lines: a red line representing a cost of £0.1 per model fit and an orange line representing a cost of £0.01 per model fit.

| Field names | Data fields | Data coding | Meaning |
|---|---|---|---|
| ***Alzheimer's disease*** | | | |
| Non-cancer Illness Codes | 20002 | 1263 | Dementia/Alzheimer's/cognitive impairment |
| Underlying (primary) cause of death: ICD10 | 40001 | F00.X, F00.0, F00.1, F00.2, F00.9; G30.X, G30.0, G30.1, G30.8, G30.9 | Dementia in Alzheimer's disease; Alzheimer's disease |
| Contributory (secondary) causes of death: ICD10 | 40002 | F00.X, F00.0, F00.1, F00.2, F00.9; G30.X, G30.0, G30.1, G30.8, G30.9 | Dementia in Alzheimer's disease; Alzheimer's disease |
| Diagnoses – ICD 10 | 41270 | F00.X, F00.0, F00.1, F00.2, F00.9; G30.X, G30.0, G30.1, G30.8, G30.9 | Dementia in Alzheimer's disease; Alzheimer's disease |
| Diagnoses – ICD 9 | 41271 | 3310 | Alzheimer's disease |
| ***Breast cancer*** | | | |
| Cancer Code | 20001 | 1002 | Breast cancer |
| Underlying (primary) cause of death: ICD10 | 40001 | C50.X, C50.0, C50.1, C50.2, C50.3, C50.4, C50.5, C50.6, C50.8, C50.9 | Malignant neoplasm of breast |
| Contributory (secondary) causes of death: ICD10 | 40002 | C50.X, C50.0, C50.1, C50.2, C50.3, C50.4, C50.5, C50.6, C50.8, C50.9 | Malignant neoplasm of breast |
| Diagnoses – ICD 10 | 41270 | C50.X, C50.0, C50.1, C50.2, C50.3, C50.4, C50.5, C50.6, C50.8, C50.9 | Malignant neoplasm of breast |
| Diagnoses – ICD 9 | 41271 | 174, 1740, 1741, 1742, 1743, 1744, 1745, 1746, 1748, 1749; 175, 1759 | Malignant neoplasm of female breast; Malignant neoplasm of male breast |
| ***Cardiovascular disease*** | | | |
| Non-cancer Illness Codes | 20002 | 1074, 1075, 1082, 1583 | Angina, Heart attack/ myocardial infarction, Transient ischaemic attack, ischaemic stroke |
| Operation code | 20004 | 1070, 1071, 1095, 1105, 1109, 1514 | Coronary angioplasty ± stent, Other arterial surgery/revascularisation procedures, Coronary artery bypass grafts, Carotid artery surgery/endarterectomy, Carotid artery angioplasty ± stent, Coronary angiogram |
| Underlying (primary) cause of death: ICD10 | 40001 | G45.X, G45.0, G45.l, G45.2, G45.3, G45.4, G45.8, G45.9; I20.X, I20.0, I20.1, I20.8, I20.9; I21.X, I21.0, I21.1, I21.2, I21.3, I21.4, I21.9; I22.X, I22.0, I22.1, I22.8, I22.9; I23.X, I23.1, I23.2, I23.3, I23.6, I23.8; I24.X, I24.0, I24.1, I24.8, I24.9; I25.X, I25.0, I25.1, I25.2, I25.3, I25.4, I25.5, I25.6, I25.8, I25.9; I63.X, I63.0, I63.1, I63.2, I63.3, I63.4, I63.5, I63.6, I63.8, I63.9; I64 | Transient cerebral ischaemic attacks and related syndromes; Angina pectoris; Acute myocardial infarction; Subsequent myocardial infarction; Certain current complications following acute myocardial infarction; Other acute ischaemic heart diseases; Chronic ischaemic heart disease; Cerebral infarction; Stroke, not specified as haemorrhage or infarction |
| Contributory (secondary) causes of death: ICD10 | 40002 | G45.X, G45.0, G45.l, G45.2, G45.3, G45.4, G45.8, G45.9; I20.X, I20.0, I20.1, I20.8, I20.9; I21.X, I21.0, I21.1, I21.2, I21.3, I21.4, I21.9; I22.X, I22.0, I22.1, I22.8, I22.9; I23.X, I23.1, I23.2, I23.3, I23.6, I23.8; I24.X, I24.0, I24.1, I24.8, I24.9; I25.X, I25.0, I25.1, I25.2, I25.3, I25.4, I25.5, I25.6, I25.8, I25.9; I63.X, I63.0, I63.1, I63.2, I63.3, I63.4, I63.5, I63.6, I63.8, I63.9; I64 | Transient cerebral ischaemic attacks and related syndromes; Angina pectoris; Acute myocardial infarction; Subsequent myocardial infarction; Certain current complications following acute myocardial infarction; Other acute ischaemic heart diseases; Chronic ischaemic heart disease; Cerebral infarction; Stroke, not specified as haemorrhage or infarction |
| Diagnoses – ICD 10 | 41270 | G45.X, G45.0, G45.l, G45.2, G45.3, G45.4, G45.8, G45.9; I20.X, I20.0, I20.1, I20.8, I20.9; I21.X, I21.0, I21.1, I21.2, I21.3, I21.4, I21.9; I22.X, I22.0, I22.1, I22.8, I22.9; I23.X, I23.1, I23.2, I23.3, I23.6, I23.8; I24.X, I24.0, I24.1, I24.8, I24.9; I25.X, I25.0, I25.1, I25.2, I25.3, I25.4, I25.5, I25.6, I25.8, I25.9; I63.X, I63.0, I63.1, I63.2, I63.3, I63.4, I63.5, I63.6, I63.8, I63.9; I64 | Transient cerebral ischaemic attacks and related syndromes; Angina pectoris; Acute myocardial infarction; Subsequent myocardial infarction; Certain current complications following acute myocardial infarction; Other acute ischaemic heart diseases; Chronic ischaemic heart disease; Cerebral infarction; Stroke, not specified as haemorrhage or infarction |
| Diagnoses – ICD 9 | 41271 | 410, 4109; 411, 4119; 412, 4129; 413, 4139; 414, 4140, 4141, 4148, 4149; 434, 4340, 4341, 4349; 436, 436; 42979 | Acute myocardial infarction; Other acute and subacute forms of ischaemic heart disease; Old myocardial infarction; Angina pectoris; Other forms of chronic ischaemic heart disease; Occlusion of cerebral arteries; Acute but ill-defined cerebrovascular disease |
| Operative procedures – OPCS4 | 41272 | K40.X, K401, K402, K403, K404, K408, K409; K41, K411, K412, K413, K414, K418, K419; K42, K421, K422, K423, K424, K428, K429; K43, K431, K432, K433, K434, K438, K439; K44, K441, K442, K448, K449; K45, K451, K452, K453, K454, K455, K456, K458, K459; K46, K461, K462, K463, K464, K465, K468, K469; K471; K49, K491, K492, K493, K494, K498, K499; K50, K501, K502, K503, K504, K508, K509; K75, K751, K752, K753, K754, K758, K759 | Saphenous vein graft replacement of coronary artery; Other autograft replacement of coronary artery; Allograft replacement of coronary artery; Prosthetic replacement of coronary artery; Other replacement of coronary artery; Connection of thoracic artery to coronary artery; Other bypass of coronary artery; Transluminal balloon angioplasty of coronary artery; Other therapeutic transluminal operations on coronary artery; Percutaneous transluminal balloon angioplasty and insertion of stent into coronary artery |

**Supplementary Table 1: Endpoint definitions:** Detailed UK Biobank field names and IDs, alongside the relevant disease codes used to extract the endpoints.

**(Available as separate file): Supplementary Table 2: Predictor matrix definitions:** Detailed UK Biobank field names and IDs, alongside the relevant data coding and preprocessing transformations.

**(Available as separate file): Supplementary Table 3: Discrimination metrices:** Detailed discrimination metrices across endpoints and predictor matrices; summarized (mean and 95% confidence intervals) across cross-validation splits.